\documentclass[10pt,twocolumn,letterpaper]{article}

\usepackage[pagenumbers]{cvpr}

\newcommand{\TODO}[1]{\textbf{\color{red}[TODO: #1]}}
\renewcommand{\TODO}[1]{}

\usepackage{multirow}
\usepackage{colortbl}

\definecolor{cvprblue}{rgb}{0.21,0.49,0.74}
\definecolor{vergerow}{HTML}{EBF5FB}
\usepackage[pagebackref,breaklinks,colorlinks,allcolors=cvprblue]{hyperref}

\newcommand{\vividurl}{https://github.com/step-out/VIVID}
\newcommand{\vividhfurl}{https://huggingface.co/ChengqianMa/VIVID}

\newcommand\blfootnote[1]{%
  \begingroup
  \renewcommand\thefootnote{}%
  \NoHyper\footnote{#1}\endNoHyper
  \addtocounter{footnote}{-1}%
  \endgroup
}

\title{VIVID: Beyond Static Multi-Image Reasoning with Localized Visual Differences and Textual Entity Dynamics in Video MLLMs}

\author{
Chengqian Ma\textsuperscript{1,$*$}\quad
Yuanze Zeng\textsuperscript{1,$*$}\quad
Hongyu Wang\textsuperscript{2}\quad
Mingfeng Lin\textsuperscript{3}\quad
YiCheng Xiao\textsuperscript{4}\quad
Yu Chen\textsuperscript{5}\\
Pengyu Li\textsuperscript{6}\quad
WenHao Wang\textsuperscript{7}\quad
Yuexiao Ma\textsuperscript{8}\quad
Xiawu Zheng\textsuperscript{9}\quad
Fei Chao\textsuperscript{9,$\dagger$}\\[2pt]
{\small
\textsuperscript{1}Peking University\quad
\textsuperscript{2}Shanghai Jiao Tong University\quad
\textsuperscript{3}Harbin Institute of Technology\quad
\textsuperscript{4}Southeast University}\\
{\small
\textsuperscript{5}Institute of Automation, Chinese Academy of Sciences\quad
\textsuperscript{6}Xi'an Jiaotong University}\\
{\small
\textsuperscript{7}Zhejiang University\quad
\textsuperscript{8}Nanyang Technological University\quad
\textsuperscript{9}Xiamen University}\\[2pt]
{\tt\small machengqian25@stu.pku.edu.cn}
}

\begin{document}
\twocolumn[{
\renewcommand\twocolumn[1][]{#1}
\vspace{-2.6em}
\maketitle
\vspace{-1.9em}
\begin{center}
  \includegraphics[width=0.70\textwidth]{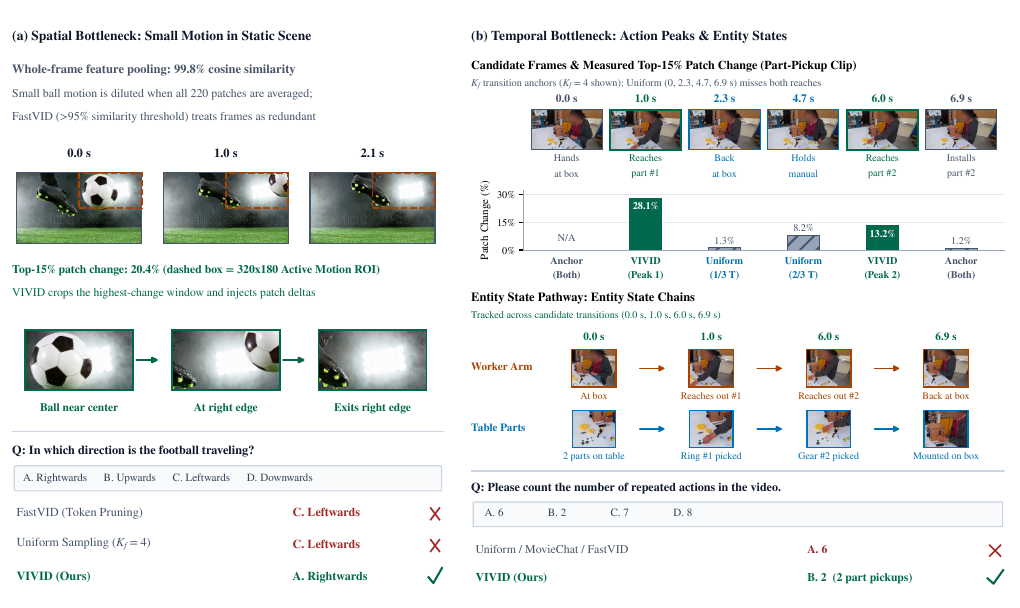}
  \vspace{-0.7em}
  \captionof{figure}{\textbf{Spatial and temporal bottlenecks of static multi image reasoning, and how VIVID resolves both} (shown at $K_f=4$ for clarity; $K_f=8$ in benchmarks):
  \textbf{(a) Spatial Bottleneck:} whole frame pooling dilutes small moving targets into static backgrounds, whereas VIVID crops the \textbf{Active Motion Region of Interest (ROI)} and folds surrounding patch differences into \textbf{Dynamic Tokens};
  \textbf{(b) Temporal Bottleneck:} equal interval sampling steps over brief action peaks, whereas \textbf{ADKP} captures both reaches and precomputed \textbf{Entity State Chains} count repeated actions accurately.}
  \label{fig:teaser}
\end{center}

\vspace{0.2em}
}]
\begin{abstract}
\blfootnote{$^*$Equal contribution.\quad $^\dagger$Corresponding author.\\Code: \url{\vividurl}\\Weights: \url{\vividhfurl}}%
Video question answering under tight frame budgets requires preserving localized motion and event order, yet current video multimodal large language models (MLLMs) reduce comprehension to static multi image reasoning by discarding unselected frames and encoding keyframes independently:
\textbf{(1) Spatially}, whole frame pooling dilutes small moving targets into static backgrounds;
\textbf{(2) Temporally}, dropping intermediate frames or averaging tokens erases inter frame motion, confusing direction, speed, and order.
We propose \textbf{VIVID} (\textbf{V}isual \textbf{I}nter frame \textbf{V}ariations and \textbf{I}ndividual \textbf{D}ynamics), which moves beyond static snapshots like human vision: keyframes capture core states while inter frame \textbf{Patch Diffs} preserve neighboring changes within $K_f$ keyframes.
\textbf{Autonomous Delta Driven Keyframe Planning (ADKP)} selects anchors from a $3\,\mathrm{fps}$ pool and feeds surrounding patch differences into \textbf{Dual Road Change Memory (DRCM)}, forming \textbf{Dynamic Tokens} at \textbf{zero extra anchor tokens} alongside a camera stabilized \textbf{Active Motion Region of Interest (ROI)} and chronological \textbf{Entity State Chains}.
Adding $0.22\%$ trainable parameters to a frozen $27\,\mathrm{B}$ backbone, \textbf{VIVID-Small} ($50$ examples, ${\sim}1.5$ GPU hours) beats all eleven baselines in overall accuracy across $8{,}083$ questions and trails only \textbf{VIVID-Large} ($3{,}600$ examples), which leads on MotionBench, Video-MME, and TempCompass.
\end{abstract}

\section{Introduction}
\label{sec:intro}

Video understanding differs fundamentally from image recognition because meaning resides in how visual entities and their states evolve across frames \cite{DBLP:journals/corr/abs-2406-07476,DBLP:journals/corr/abs-2409-12191,DBLP:journals/corr/abs-2502-13923,DBLP:journals/tmlr/0080ZGZ00ZZL0L25}.
However, current video multimodal large language models (MLLMs) largely reduce video comprehension to \textbf{static multi image reasoning}: their vision encoders process each frame independently as a static image \cite{DBLP:conf/eccv/ChenZLBLZC24,DBLP:conf/cvpr/YangCTWL0J25,DBLP:conf/cvpr/TaoQYSW25,DBLP:conf/acl/HuangZH25,DBLP:conf/nips/ShaoTQYSW25,shen2025fastvid,wang2026earlytom,luo2026stsimdiff} or condense static features into compressed memory and summary representations \cite{DBLP:conf/cvpr/SongCWZZWCG0ZLH24,DBLP:conf/cvpr/0004LJJCSSL24,DBLP:conf/icml/ShenX0W0ZLXVBL025,DBLP:conf/cvpr/ShuLZQ0L0025,DBLP:conf/nips/QianDZZDLW24,liu2026videomind,minh2026protokv,lu2026vista,fu2026videostir}, forcing the language model to guess inter frame transitions implicitly.
As shown in \cref{fig:teaser}, this static paradigm suffers from dual bottlenecks in \textbf{space} and \textbf{time}:

\paragraph{Spatial Bottleneck (Similarity Based Token Compression and Frame Selection).}
Token compression and frame selection methods \cite{DBLP:conf/eccv/ChenZLBLZC24,DBLP:conf/cvpr/TaoQYSW25,DBLP:conf/acl/HuangZH25,DBLP:conf/nips/ShaoTQYSW25,shen2025fastvid,luo2026stsimdiff,wang2026earlytom} prune redundancy via attention scores, token clustering, or whole frame similarity, yet critical motions often span only a small image region against a static background.
In \cref{fig:teaser}(a), as the soccer ball moves rightward through the upper frame, mean pooled Vision Transformer (ViT) cosine similarity between $0.0\,\mathrm{s}$ and $1.0\,\mathrm{s}$ remains $99.8\%$ ($0.9983$): under localized perturbations (\cref{sec:method_prelim}), global cosine distance scales quadratically with changed patch fraction $\rho$, submerging small moving targets, whereas averaging the top $15\%$ most changed patches yields a $20.4\%$ local signal.

\paragraph{Temporal Bottleneck (Long Horizon Video Memory and Event Grounding).}
Long horizon memory and event grounding methods \cite{DBLP:conf/cvpr/SongCWZZWCG0ZLH24,DBLP:conf/cvpr/0004LJJCSSL24,DBLP:conf/icml/ShenX0W0ZLXVBL025,DBLP:conf/cvpr/ShuLZQ0L0025,DBLP:conf/nips/QianDZZDLW24,Chen_2024_CVPR,liu2026videomind,minh2026protokv} average adjacent frame features into memory slots or summarize segments into coarse captions.
Equal interval sampling steps over brief transition peaks (such as $1.0\,\mathrm{s}$ and $6.0\,\mathrm{s}$ in \cref{fig:teaser}(b)), while timeless feature pooling and clip captions entangle multiple objects and timestamps, confusing direction, speed, repetition count, and order \cite{DBLP:conf/acl/LiuLLWRLCSH24,DBLP:conf/iclr/ShangguanLDZ0FC25,Hong_2025_CVPR}.

\begin{figure*}[!t]
  \centering
  \includegraphics[width=0.98\textwidth]{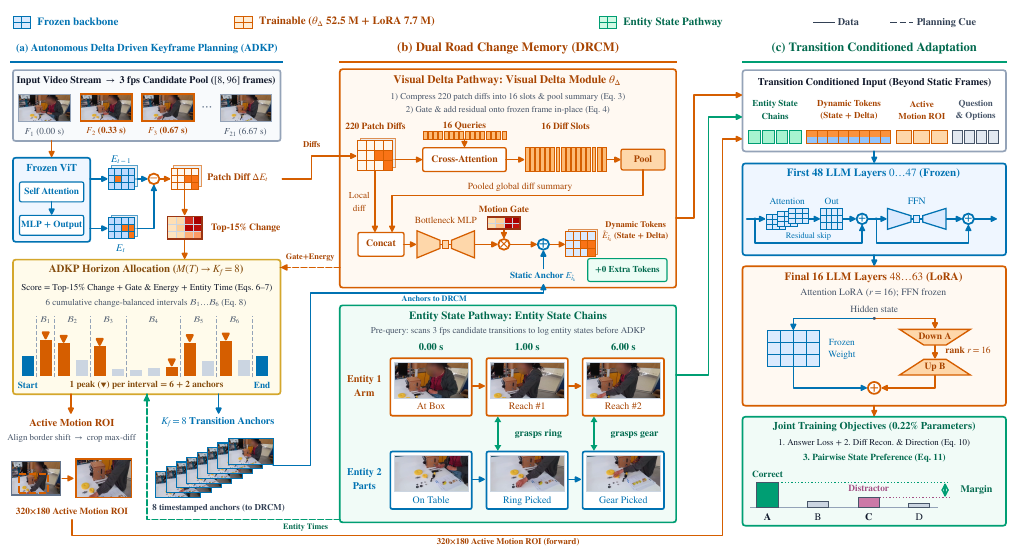}
  \caption{\textbf{Overview of the VIVID architecture} (ViT: Vision Transformer; MLP: multilayer perceptron; FFN: feedforward network; ROI: region of interest; LoRA: Low Rank Adaptation):
  \textbf{(a)} across $3\,\mathrm{fps}$ candidates (\cref{sec:method}), the frozen ViT computes \textbf{Patch Diffs} $\Delta E_t$ so \textbf{ADKP} (\cref{eq:loc_and_energy,eq:plan_score,eq:bucket_select}) selects $K_f=8$ transition anchors across intervals $\mathcal{B}_1\dots\mathcal{B}_6$ and crops a $320\times180$ \textbf{Active Motion ROI};
  \textbf{(b)} surrounding \textbf{Patch Diffs} enter \textbf{Dual Road Change Memory (DRCM)}, where the Visual Delta Module $\theta_\Delta$ enriches static anchors $E_{\hat{t}_k}$ into \textbf{Dynamic Tokens} $\tilde{E}_{\hat{t}_k}$ (\cref{eq:delta_resampler,eq:gated_enhancer}) alongside precomputed \textbf{Entity State Chains} (\cref{eq:entity_chain});
  \textbf{(c)} the \textbf{Transition Conditioned Input} feeds the shared reader, with the first $48$ LLM layers frozen and the final $16$ layers adapted via LoRA.}
  \label{fig:architecture}
\end{figure*}

\paragraph{Our Paradigm Shift: VIVID.}
Conventional keyframe selection discards unselected frames and feeds chosen keyframes to the LLM as isolated pictures, blind to adjacent frame motion.
By contrast, human vision pairs sparse fixations on key states with sensitivity to inter frame differences.
Inspired by this principle, we propose \textbf{VIVID} (\textbf{V}isual \textbf{I}nter frame \textbf{V}ariations and \textbf{I}ndividual \textbf{D}ynamics; \cref{fig:architecture}): although the reader processes only $K_f$ keyframes, VIVID uses inter frame \textbf{Patch Diffs} ($\Delta E_t = E_t - E_{t-1}$, where $E_t$ denotes the patch feature matrix of candidate frame $t$) both to plan keyframes via \textbf{Autonomous Delta Driven Keyframe Planning (ADKP, \cref{sec:method_adkp})} and to inject neighboring frame motion into \textbf{Dual Road Change Memory (DRCM, \cref{sec:method_drcm})} inside the \textbf{Transition Conditioned Input}:
(1) in the \textbf{Visual Delta Pathway}, our trainable \textbf{Visual Delta Module} ($\theta_\Delta$) compresses $220$ patch differences into $16$ latent slots and injects gated motion residuals into each selected keyframe to form \textbf{Dynamic Tokens} ($\tilde{E}_t$, encoding both keyframe appearance and neighboring frame changes) at \textbf{zero extra anchor tokens}, paired with a camera stabilized \textbf{Active Motion Region of Interest (ROI)}; and
(2) in the \textbf{Entity State Pathway}, a vision language writer tracks candidate transitions into chronological \textbf{Entity State Chains}.
Thus, neighboring differences and entity observations beyond the $K_f$ keyframes are retained rather than discarded.
Zero initialization ($\tilde{E}_t=E_t$ at step $0$) and dense $220$ patch reconstruction allow joint adaptation of $\theta_\Delta$ and Low Rank Adaptation (\textbf{LoRA}) in the final $16$ layers ($\theta_{\mathrm{LoRA}}$, $0.22\%$ of the $27\,\mathrm{B}$ backbone; \cref{sec:method_lora}) to deliver strong gains under minimal resources (\textbf{VIVID-Small}, $50$ examples in ${\sim}1.5$ H20 GPU hours) and scale cleanly on a compact $3{,}600$ clip corpus (\textbf{VIVID-Large}, $<0.8\%$ of the $481\,\mathrm{K}$ training samples in VideoMind \cite{liu2026videomind}).

Our main contributions are threefold:
\begin{itemize}
    \item \textbf{Surrounding frame Dynamic Tokens:} Rather than discarding unselected frames, our \textbf{Visual Delta Module} ($\theta_\Delta$) folds neighboring frame \textbf{Patch Diffs} into each keyframe as \textbf{Dynamic Tokens} ($\tilde{E}_t$) at zero extra anchor token cost (\cref{sec:method_drcm}).
    \item \textbf{Dual Road Change Memory and ADKP:} We unify ADKP planning with Dynamic Tokens, camera stabilized \textbf{Active Motion ROI}, and chronological \textbf{Entity State Chains} across \textbf{VIVID-Small} ($50$) and \textbf{VIVID-Large} ($3{,}600$; \cref{sec:method_adkp,sec:method_lora}).
    \item \textbf{Controlled multi benchmark verification:} Under a strictly matched $K_f=8$ protocol on a shared $27\,\mathrm{B}$ backbone across MotionBench \cite{Hong_2025_CVPR}, Video-MME \cite{DBLP:conf/cvpr/FuDLLRZWZSZCLLZ25}, and TempCompass \cite{DBLP:conf/acl/LiuLLWRLCSH24} ($8{,}083$ questions), VIVID outperforms ten published baselines and uniform sampling across token compression, frame selection, memory, and grounding (\cref{sec:experiments}).
\end{itemize}

\section{Related Work}
\label{sec:related}

\paragraph{Similarity Based Token Compression and Frame Selection.}
Early ViT pruning merges similar tokens (ToMe \cite{DBLP:conf/iclr/BolyaFDZFH23}, vid-TLDR \cite{DBLP:conf/cvpr/ChoiLCCK24}) or drops low attention tokens (FastV \cite{DBLP:conf/eccv/ChenZLBLZC24}, VisionZip \cite{DBLP:conf/cvpr/YangCTWL0J25}, LLaVA-PruMerge \cite{DBLP:conf/iccv/ShangCXLY25}).
Recent video MLLMs extend reduction across frames: DyCoke \cite{DBLP:conf/cvpr/TaoQYSW25} merges temporal groups, PruneVid \cite{DBLP:conf/acl/HuangZH25} clusters static and dynamic tokens, HoliTom \cite{DBLP:conf/nips/ShaoTQYSW25} performs spatio temporal merging, FastVID \cite{shen2025fastvid} segments clips by whole frame similarity, EarlyTom \cite{wang2026earlytom} compresses spatial tokens early, and ST-SimDiff \cite{luo2026stsimdiff} balances similarity prototypes with temporal difference token selection.
Complementing selection and pruning, VIVID learns to inject surrounding \textbf{Patch Diffs} directly into anchor embeddings as \textbf{Dynamic Tokens} while guiding ADKP with learned gates and top $15\%$ local changes.

\paragraph{Long Horizon Video Memory and Event Grounding.}
To scale video MLLMs to multi minute benchmarks \cite{DBLP:conf/nips/MangalamAM23,DBLP:conf/nips/WuLCL24,DBLP:conf/cvpr/0001S0WLXQYXZH025}, long video methods condense streams into feature slots or text summaries: MovieChat \cite{DBLP:conf/cvpr/SongCWZZWCG0ZLH24}, MA-LMM \cite{DBLP:conf/cvpr/0004LJJCSSL24}, Chat-UniVi \cite{DBLP:conf/cvpr/0001TZC024}, LLaMA-VID \cite{DBLP:conf/eccv/LiWJ24}, LongVU \cite{DBLP:conf/icml/ShenX0W0ZLXVBL025}, Video-XL \cite{DBLP:conf/cvpr/ShuLZQ0L0025}, Oryx \cite{DBLP:conf/iclr/LiuD0HLR25}, ProtoKV \cite{minh2026protokv}, and Vista \cite{lu2026vista} pool historical frames into compressed slots or prototypes, whereas VideoLLM-online \cite{Chen_2024_CVPR}, VideoStreaming \cite{DBLP:conf/nips/QianDZZDLW24}, Goldfish \cite{DBLP:conf/eccv/AtaallahSASZDZSE24}, VideoChat-Online \cite{Huang_2025_CVPR_OVBench}, VideoMind \cite{liu2026videomind}, and VideoStir \cite{fu2026videostir} generate clip captions or retrieve events.
To preserve fine grained per object order alongside compact memory, VIVID pairs \textbf{Dynamic Tokens} with structured per entity \textbf{Entity State Chains}.

\paragraph{Overcoming Static Shortcut Bias in Video Temporal Adaptation.}
Prior diagnostic studies \cite{DBLP:conf/cvpr/BuchEG00N22,DBLP:conf/acl/0003BB23} demonstrate that video language models frequently rely on single frame appearance cues rather than inter frame transitions, motivating dedicated temporal and motion benchmarks \cite{DBLP:conf/cvpr/0002WH00LWX0L0024,DBLP:conf/acl/LiuLLWRLCSH24,DBLP:conf/eccv/LiLLRLGSH24,cores2024tvbench,DBLP:conf/iclr/ShangguanLDZ0FC25,Hong_2025_CVPR}.
Whereas LoRA \cite{DBLP:conf/iclr/HuSWALWWC22} provides parameter efficient adaptation and Direct Preference Optimization (DPO) \cite{DBLP:conf/nips/RafailovSMMEF23} incorporates a reference policy likelihood term, VIVID supervises adaptation with dense patch difference reconstruction and a reference free option preference margin loss.

\section{Methodology}
\label{sec:method}

Given a video of duration $T$ seconds, \textbf{VIVID} samples a candidate pool of $M(T)=\mathrm{clip}(\lfloor 3.0\,T\rfloor+1, 8, 96)$ frames $F_1,\dots,F_M$ at timestamps $\tau_1<\dots<\tau_M$, where $\mathrm{clip}(\cdot,8,96)$ bounds candidate count to $[8,96]$ across short clips and long videos. Analogous to human vision, VIVID folds neighboring frame patch differences into $K_f=8$ selected keyframes while keeping all $27\,\mathrm{B}$ backbone weights frozen (\cref{fig:architecture}):
the frozen encoder $\Phi_{\mathrm{vis}}$ extracts patch matrices $E_t = \Phi_{\mathrm{vis}}(F_t) \in \mathbb{R}^{N_p\times D}$, where $N_p=220$ is the patch count and $D=5120$ is the embedding width (\cref{sec:method_prelim}); \textbf{DRCM} (\cref{sec:method_drcm}) builds \textbf{Dynamic Tokens}, an \textbf{Active Motion ROI}, and \textbf{Entity State Chains}; \textbf{ADKP} (\cref{sec:method_adkp}) selects $K_f=8$ transition anchors; and \textbf{Joint Adaptation} (\cref{sec:method_lora}) optimizes $\theta_\Delta$ with LoRA in the final $16$ layers ($\theta_{\mathrm{LoRA}}$) and a learned pathway gate.
Throughout, $X^{(i)} \in \mathbb{R}^d$ denotes the transposed $i$th row of $X\in\mathbb{R}^{N\times d}$.

\subsection{Preliminaries: Attenuation of Localized Changes}
\label{sec:method_prelim}

Whole frame similarity methods summarize frame $F_t$ ($20\times 11$ grid of $32\times 32$ merged patch cells, $N_p=220$) by its unit normalized mean patch vector $g_t = \bar{E}_t / \max(\|\bar{E}_t\|_2, \epsilon)$ (where $\bar{E}_t = \frac{1}{N_p}\sum_{i=1}^{N_p} E_t^{(i)}$ and $\epsilon=10^{-12}$; \cref{eq:prelim_encode}) and treat consecutive frames as redundant when $s_{\mathrm{global}} = g_{t-1}^\top g_t$ approaches unity \cite{shen2025fastvid}.
Suppose only a fraction $\rho\in(0,1)$ of patch embeddings shift with mean $d_t\in\mathbb{R}^D$ while the remaining $1-\rho$ embeddings stay unchanged ($\bar{E}_t = \bar{E}_{t-1} + \rho d_t$).
For small localized perturbations ($\rho\|d_t\|_2 \ll \|\bar{E}_{t-1}\|_2$), unit normalization cancels the first order radial component along $g_{t-1}$, so a second order Taylor expansion yields \cref{eq:prelim_taylor}:
\begin{equation}
    1 - s_{\mathrm{global}} \approx \frac{\rho^2\|d_{t,\perp}\|_2^2}{2\|\bar{E}_{t-1}\|_2^2},
\label{eq:prelim_taylor}
\end{equation}
where $d_{t,\perp} = d_t - (g_{t-1}^\top d_t)g_{t-1}$ is the component orthogonal to $\bar{E}_{t-1}$.
Because the leading term scales with $\rho^2$, small moving targets ($\rho \ll 1$) are attenuated by static backgrounds: in \cref{fig:teaser}(a), $s_{\mathrm{global}} = 0.9983$ exceeds FastVID's $0.95$ threshold despite a $20.4\%$ cosine change across the top $15\%$ patches.

\subsection{Dual Road Change Memory (DRCM)}
\label{sec:method_drcm}

\subsubsection{Visual Delta Pathway: Visual Delta Module and Active Motion ROI}
To enrich frozen visual snapshots into \textbf{Dynamic Tokens} with explicit inter frame dynamics at zero extra anchor tokens, our trainable \textbf{Visual Delta Module} ($\theta_\Delta$, $52.5\,\mathrm{M}$ parameters) processes consecutive candidate pairs $(F_{t-1},F_t)$ during planning and neighboring same scene reference pairs $(F_{\mathrm{ref}},F_{\hat{t}_k})$ during reading (leaving anchor $1$ unchanged; \cref{sec:supp_impl}).
For any frame pair with current embedding $E_t$ and reference $E_{t-1}$, we compute the patch difference matrix $\Delta E_t$ and nonnegative cosine distance $m_t^{(i)}$ at each patch $i\in\{1,\dots,N_p\}$:
\begin{equation}
\left\{
\begin{aligned}
    \Delta E_t &= E_t - E_{t-1} \in \mathbb{R}^{N_p \times D}, \\
    m_t^{(i)} &= \max\!\left(0, \; 1 - \hat{E}_{t-1}^{(i)\top} \hat{E}_t^{(i)}\right),
\end{aligned}
\right.
\label{eq:patch_mag}
\end{equation}
where $\hat{E}_t^{(i)} = E_t^{(i)} / \max(\|E_t^{(i)}\|_2, \epsilon)$ denotes the unit normalized patch vector.
\paragraph{Cross Attention Difference Resampler.}
Let $\tilde{D}_t = \mathrm{LN}_\Delta(\Delta E_t)$ denote layer normalized patch differences.
To prevent static background patches from dominating attention, we weight each patch by $(0.2 + 5.0\,m_t^{(i)})$ and compress all $N_p=220$ patch differences into $K=16$ latent slots $Z_t\in\mathbb{R}^{K\times D}$ via bottleneck cross attention ($d_b=512$) with learned queries $Q_\Delta\in\mathbb{R}^{K\times d_b}$:
\begin{equation}
\left\{
\begin{aligned}
    \widehat{\Delta E}_t^{(i)} &= \big(0.2 + 5.0 \, m_t^{(i)}\big) \, \tilde{D}_t^{(i)}, \\
    A_t &= \mathrm{softmax}\!\left(\frac{Q_\Delta (\widehat{\Delta E}_t W_K)^\top}{\sqrt{d_b}}\right), \\
    Z_t &= \mathrm{SiLU}\!\left(A_t \widehat{\Delta E}_t W_V\right) W_O,
\end{aligned}
\right.
\label{eq:delta_resampler}
\end{equation}
where $A_t\in\mathbb{R}^{K\times N_p}$ is the patch normalized attention matrix, $\mathrm{SiLU}$ is the sigmoid linear unit, and $W_K,W_V\in\mathbb{R}^{D\times d_b}$ and $W_O\in\mathbb{R}^{d_b\times D}$ are learned projections.
\paragraph{Zero Overhead Gated Residual Injection.}
Rather than appending extra tokens, a bottleneck multilayer perceptron (MLP) combines each patch's local difference $\tilde{D}_t^{(i)}$ with the pooled summary $\bar{z}_t = \frac{1}{K}\sum_{k=1}^K Z_t^{(k)}$ and injects a gated residual into $E_t^{(i)}$ in place:
\begin{equation}
\left\{
\begin{aligned}
    G_t^{(i)} &= \sigma\!\left(w_g^\top \tilde{D}_t^{(i)} + b_g + 2.0 \, m_t^{(i)}\right), \\
    h_t^{(i)} &= \mathrm{SiLU}\!\left(W_{\mathrm{down}}\big[\tilde{D}_t^{(i)}\,;\,\bar{z}_t\big] + b_{\mathrm{down}}\right), \\
    U_t^{(i)} &= W_{\mathrm{up}}h_t^{(i)}+b_{\mathrm{up}}, \\
    \tilde{E}_t^{(i)} &= E_t^{(i)} + \tanh(\alpha_{\mathrm{res}}) \, G_t^{(i)} \, U_t^{(i)},
\end{aligned}
\right.
\label{eq:gated_enhancer}
\end{equation}
where $\sigma$ is the logistic sigmoid, $[\,\cdot\,;\,\cdot\,]$ denotes vector concatenation, $G_t^{(i)}\in(0,1)$ is the learned motion gate parameterized by $w_g\in\mathbb{R}^D$ and $b_g\in\mathbb{R}$, $h_t^{(i)}\in\mathbb{R}^{d_b}$ and $U_t^{(i)}\in\mathbb{R}^D$ are hidden and residual vectors parameterized by $W_{\mathrm{down}}\in\mathbb{R}^{d_b\times 2D}, b_{\mathrm{down}}\in\mathbb{R}^{d_b}, W_{\mathrm{up}}\in\mathbb{R}^{D\times d_b}, b_{\mathrm{up}}\in\mathbb{R}^D$, and scalar $\alpha_{\mathrm{res}}$ (initialized to $0.1$) controls residual scale.
Initializing $W_O$, $W_{\mathrm{up}}$, and $b_{\mathrm{up}}$ to zero ensures $\tilde{E}_t = E_t$ at step $0$, preserving pretrained alignment at \textbf{zero extra anchor tokens}.
To resolve tiny targets spanning only a few $32\times 32$ cells, we estimate 2D patch shifts $(\Delta x_t,\Delta y_t)$ across the top $25\%$ changed patches within a $(2r+1)\times(2r+1)$ grid window ($r=3$), align the outer $20\%$ image border within $[-6,6]$ pixels on $160\times 90$ grayscale frames to compensate camera translation, and crop the highest residual $80\times 45$ window at $320\times 180$ resolution as one $66$ token \textbf{Active Motion ROI} when $\mu_t^{\mathrm{loc}}\ge 0.015$ (\cref{sec:supp_impl}).

\subsubsection{Entity State Pathway: Entity State Chains}
To prevent multi frame merging from confusing entity order, speed, and interactions, the Entity State Pathway constructs chronological \textbf{Entity State Chains} across $L+1\in\{4,6\}$ selected candidate frames $\{F_{t_0},\dots,F_{t_L}\}$ defining $L$ transitions across $[0,T]$ (\cref{fig:architecture}(b)).
Before invoking the vision language writer on $(F_{t_{l-1}}, F_{t_l})$, we retrieve up to eight active entities up to $\tau_{t_{l-1}}$ into working memory (canonical name, category, attributes, and latest position/state).
Conditioned on working memory, $(F_{t_{l-1}}, F_{t_l})$, the \textbf{Active Motion ROI}, and $(\Delta x_{t_l},\Delta y_{t_l})$, the writer reuses canonical names for continuing objects, introduces new names only for unseen objects (\cref{sec:supp_prompts}), and updates interval descriptions $\mathcal{M}_{\mathrm{seg}}$ and per entity \textbf{Entity State Chains}:
\begin{equation}
\begin{aligned}
\mathcal{C}(e_j) = \big\{&(\tau_{j,0},s_{j,0},p_{j,0},\delta_{j,0})\to\dots\\
                        &\to(\tau_{j,L_j},s_{j,L_j},p_{j,L_j},\delta_{j,L_j})\big\},
\end{aligned}
\label{eq:entity_chain}
\end{equation}
where for entity $e_j$ across its $L_j+1$ logged observations ($r=0,\dots,L_j$), $\tau_{j,r}$ is the timestamp, $s_{j,r}$ and $p_{j,r}$ denote state and screen position, and $\delta_{j,r}$ logs state change, direction, speed, and inter entity relations.
For question $Q$, retrieval selects up to six transitions per relevant entity and ranks candidate intervals by content word overlap with $Q$ plus local patch salience $\mu_t^{\mathrm{loc}}$, preserving chronological order up to the query timestamp (\cref{sec:supp_prompts}).

\subsection{Autonomous Delta Driven Keyframe Planning (ADKP)}
\label{sec:method_adkp}

Unlike static selection methods that discard unselected frames, VIVID reuses candidate \textbf{Patch Diffs} both to select $K_f=8$ keyframes here and to form neighboring frame \textbf{Dynamic Tokens} $\tilde{E}_{\hat{t}_k}$ inside DRCM. ADKP captures active state changes without clustering slots in a single burst using three statistics from \cref{eq:patch_mag,eq:delta_resampler,eq:gated_enhancer}:
\begin{equation}
\left\{
\begin{aligned}
    \mu_t^{\mathrm{gate}} &= \frac{1}{k_{20}} \sum_{i \in \mathcal{I}_{0.20}(G_t)} G_t^{(i)}, \\
    \mu_t^{\mathrm{loc}} &= \frac{1}{k_{15}} \sum_{i \in \mathcal{I}_{0.15}(m_t)} m_t^{(i)}, \\
    \varepsilon_t^\Delta &= \frac{\|Z_t\|_F}{\sqrt{K D}},
\end{aligned}
\right.
\label{eq:loc_and_energy}
\end{equation}
where $\mathcal{I}_p(v)$ indexes the largest $p N_p$ entries of $v$ ($k_{20}=44$ for \textbf{gate salience} $\mu_t^{\mathrm{gate}}$; $k_{15}=33$ for \textbf{local change} $\mu_t^{\mathrm{loc}}$) and $\|\cdot\|_F$ is the Frobenius norm for \textbf{slot energy} $\varepsilon_t^\Delta$.
Combining \cref{eq:loc_and_energy} with \textbf{entity timestamp indicator} $I_t^{\mathrm{mem}}(Q) = \mathbb{I}(\min_{\tau^*\in\mathcal{T}_{\mathrm{mem}}(Q)}|\tau_t-\tau^*|\le \delta_\tau)$ (where $\mathbb{I}$ is the indicator, $\mathcal{T}_{\mathrm{mem}}(Q)$ is the retrieved timestamp set with $I_t^{\mathrm{mem}}=0$ if empty, and $\delta_\tau=\max(1.25\,\mathrm{s},1.25\,\overline{\Delta\tau})$ for mean step $\overline{\Delta\tau}$, evaluating $\mu_t^{\mathrm{loc}}$ at step $1$ and $1\,\mathrm{s}$ stride $3$), the planning score for $t\in\{2,\dots,M\}$ is:
\begin{equation}
\begin{aligned}
    S_{\mathrm{plan}}(t \mid Q) &= \mu_t^{\mathrm{gate}} + \gamma_1 \mu_t^{\mathrm{loc}} \\
    &\quad + \gamma_2 \varepsilon_t^\Delta + \gamma_3 I_t^{\mathrm{mem}}(Q),
\end{aligned}
\label{eq:plan_score}
\end{equation}
where $(\gamma_1,\gamma_2,\gamma_3)=(2.5,0.5,0.35)$ are scalar weights.
When $M>K_f$, ADKP fixes endpoints $\mathcal{S}_0=\{1,M\}$, splits interior cumulative weight into $K_f-2$ intervals $\mathcal{B}_b$, and picks the peak in each nonempty interval (\cref{sec:supp_impl}):
\begin{equation}
\left\{
\begin{aligned}
    w_t &= 0.45 \, \bar{S} + 0.55 \, \tilde{S}_t, \\
    C_t &= \frac{\sum_{j=2}^t w_j}{\sum_{j=2}^{M-1} w_j}, \\
    t_b^* &= \arg\max_{t \in \mathcal{B}_b \setminus \mathcal{S}_{b-1}} S_{\mathrm{plan}}(t \mid Q),
\end{aligned}
\right.
\label{eq:bucket_select}
\end{equation}
where $\tilde{S}_t = \max(10^{-4}, S_{\mathrm{plan}}(t\mid Q))$ ($t=2,\dots,M-1$) has mean $\bar{S}$, $\mathcal{S}_b = \mathcal{S}_{b-1}\cup\{t_b^*\}$ (filling empty intervals by top score), and sorting $\mathcal{S}_{K_f-2}$ yields anchors $\hat{t}_1<\dots<\hat{t}_{K_f}$.
\paragraph{Handling Shot Cuts, Camera Pans, and Surrounding Frames.}
Ranking patch changes directly can waste anchor slots on hard shot cuts or disturb uniform speed cues during global camera pans.
ADKP therefore multiplies candidate scores by $0.1$ when whole frame similarity $g_{t-1}^\top g_t < 0.80$ or median patch distance $\mathrm{med}(m_t)>0.22$, and keeps uniform anchor spacing when global camera motion dominates ($\mu_t^{\mathrm{loc}}/(\mathrm{med}(m_t)+0.02) < 2.35$ across $75\%$ of frames).
During reading, when consecutive anchors $(\hat{t}_{k-1},\hat{t}_k)$ span a shot cut or exceed the uniform step, $\theta_\Delta$ evaluates \cref{eq:patch_mag,eq:delta_resampler,eq:gated_enhancer} on $\Delta E_{\hat{t}_k}^{\mathrm{read}} = E_{\hat{t}_k} - E_{\mathrm{ref}}$ using a neighboring candidate reference $E_{\mathrm{ref}}$ passing the scene similarity threshold ($g_{\mathrm{ref}}^\top g_{\hat{t}_k}\ge 0.82$), folding unselected neighboring frames into $\tilde{E}_{\hat{t}_k}$ at zero extra anchor tokens.

\begin{table*}[!t]
  \centering
  \scriptsize
  \setlength{\tabcolsep}{3.6pt}
  \renewcommand{\arraystretch}{1.05}
  \caption{\textbf{Per category accuracy (\%) across the six official motion categories of MotionBench} on the shared Qwen3.8 27B backbone under the unified dynamic $3\,\mathrm{fps}$ candidate pool ($M(T)\in[8,96]\to K_f=8$ reader frames; \textbf{bold} indicates best and \underline{underline} second best).}
  \label{tab:main_detailed}
  \resizebox{\textwidth}{!}{%
  \begin{tabular}{@{}llccccccc@{}}
    \toprule
    \textbf{Method} & \textbf{Venue} & \textbf{Motion Recognition} & \textbf{Moving Objects} & \textbf{Location Motion} & \textbf{Action Order} & \textbf{Repetition Count} & \textbf{Camera Motion} & \textbf{Overall} \\
    \midrule
    \multicolumn{9}{l}{\textit{Static Multi Keyframe Baseline}} \\
    Uniform Sampling & -- & 62.86 & 72.03 & 64.10 & \underline{44.59} & 27.75 & \textbf{63.90} & 58.85 \\
    \midrule
    \multicolumn{9}{l}{\textit{Similarity Based Token Compression and Frame Selection}} \\
    DyCoke \cite{DBLP:conf/cvpr/TaoQYSW25} & CVPR'25 & 63.67 & 70.29 & 64.65 & 42.28 & 29.50 & 60.52 & 58.48 \\
    PruneVid \cite{DBLP:conf/acl/HuangZH25} & ACL'25 & 56.22 & 66.23 & 54.76 & 42.47 & 24.25 & 36.62 & 50.91 \\
    HoliTom \cite{DBLP:conf/nips/ShaoTQYSW25} & NeurIPS'25 & 63.87 & 70.00 & 63.55 & 42.66 & 27.50 & 62.86 & 58.43 \\
    FastVID \cite{shen2025fastvid} & NeurIPS'25 & 62.45 & 69.71 & 64.10 & 43.05 & 28.25 & 62.34 & 58.00 \\
    ST-SimDiff \cite{luo2026stsimdiff} & ICLR'26 & 61.71 & 72.17 & 61.72 & 40.54 & 28.75 & 56.88 & 57.03 \\
    EarlyTom \cite{wang2026earlytom} & CVPR'26 & 64.21 & 71.88 & 63.37 & 41.51 & 27.75 & 62.60 & 58.70 \\
    \midrule
    \multicolumn{9}{l}{\textit{Long Horizon Video Memory and Event Grounding}} \\
    MovieChat \cite{DBLP:conf/cvpr/SongCWZZWCG0ZLH24} & CVPR'24 & 63.73 & 71.88 & 63.74 & 42.08 & 29.75 & \underline{63.64} & 58.95 \\
    VideoStreaming \cite{DBLP:conf/nips/QianDZZDLW24} & NeurIPS'24 & 65.70 & 75.51 & 62.27 & 43.82 & 28.50 & 60.78 & 59.92 \\
    VideoMind \cite{liu2026videomind} & ICLR'26 & 59.40 & 68.70 & 59.16 & 40.73 & 27.50 & 58.18 & 55.27 \\
    ProtoKV \cite{minh2026protokv} & ICML'26 & 64.48 & 71.16 & 64.10 & 43.24 & 29.75 & 62.86 & 59.22 \\
    \midrule
    \multicolumn{9}{l}{\textit{Dual Road Change Memory and Delta Driven Planning (Ours)}} \\
    \rowcolor{vergerow}
    \textbf{VIVID-Small} & \textbf{Ours} & \underline{66.64} & \underline{76.23} & \underline{67.95} & \textbf{45.37} & \textbf{33.00} & 59.74 & \underline{61.71} \\
    \rowcolor{vergerow}
    \textbf{VIVID-Large} & \textbf{Ours} & \textbf{67.93} & \textbf{77.83} & \textbf{69.23} & \underline{44.59} & \underline{32.25} & 63.12 & \textbf{62.78} \\
    \bottomrule
  \end{tabular}%
  }
\end{table*}

\subsection{Joint Adaptation and Neural Pathway Gating}
\label{sec:method_lora}

At inference, each reader context concatenates all $K_f=8$ anchors chronologically ($220$ \textbf{Dynamic Tokens} $\tilde{E}_{\hat{t}_k}$ per anchor, plus one $66$ token \textbf{Active Motion ROI} $E_{\mathrm{ROI}}$ when triggered), each preceded by an explicit timestamp header (\texttt{Frame \#k/8 at t=...s}) so the reader aligns ADKP anchors with \textbf{Entity State Chains}. Evaluating the visual context $[\{(\tau_{\hat{t}_k},\tilde{E}_{\hat{t}_k})\}\Vert Q]$ and dual road context $\mathcal{X}=[\mathcal{M}_{\mathrm{seg}}\Vert\{\mathcal{C}(e_j)\}_{j=1}^J\Vert\{(\tau_{\hat{t}_k},\tilde{E}_{\hat{t}_k})\}\Vert(\tau^*,E_{\mathrm{ROI}})\Vert Q]$ on the shared reader yields option log probabilities $\ell_{\mathrm{vis}},\ell_{\mathrm{mem}}\in\mathbb{R}^4$.
Because spatial attribute questions favor direct Dynamic Token inspection whereas multi step order and speed questions benefit more from Entity State Chains, we fuse the pathways as $(1-\lambda_\psi)\ell_{\mathrm{vis}} + \lambda_\psi\ell_{\mathrm{mem}}$ via a two layer \textbf{Neural Pathway Gate} $\lambda_\psi(h_Q, c)\in(0,1)$ trained on held out validation clips without dataset or task labels, mapping the frozen backbone's pooled middle and final layer question vector $h_Q\in\mathbb{R}^{2D}$ and six pathway predictive entropy, top $2$ margin, symmetric Kullback Leibler (KL) divergence, and prediction agreement features $c\in\mathbb{R}^6$ (zero initialized around $\lambda_\psi=0.5$; \cref{sec:supp_impl}).
To adapt the reader to decode $\tilde{E}_t$ and $\{\mathcal{C}(e_j)\}$, we keep its first $48$ layers ($0\text{--}47$, including all feedforward networks [FFNs]) frozen and attach $r=16$ attention LoRA modules ($\theta_{\mathrm{LoRA}}$, $7.7\,\mathrm{M}$ parameters) to its final $16$ layers ($48\text{--}63$; $60.2\,\mathrm{M}$ total trainable parameters, or $0.22\%$ of the $27\,\mathrm{B}$ backbone), jointly optimizing $\theta=\{\theta_\Delta,\theta_{\mathrm{LoRA}}\}$.
Beyond supervised fine tuning $\mathcal{L}_{\mathrm{SFT}} = -\log P_\theta(y^+\mid\mathcal{X})$ on ground truth option $y^+$, two auxiliary objectives guide $\theta$ from small sample counts:

\paragraph{Difference Reconstruction and Direction Alignment.}
If $\theta_\Delta$ were supervised solely by answer loss, $Z_t$ could overlook subtle spatial changes whenever coarse context suffices.
We attach an auxiliary training time decoder (discarded at inference) in which normalized preceding tokens $\mathrm{LN}_{\mathrm{ctx}}(E_{t-1})$ query $Z_t$ to reconstruct all $N_p=220$ patch differences:
\begin{equation}
\left\{
\begin{aligned}
    Q_{\mathrm{rec}} &= \mathrm{LN}_{\mathrm{ctx}}(E_{t-1})W_Q^{\mathrm{rec}}, \\
    K_{\mathrm{rec}} &= (Z_t+Q_\Delta W_K^\top)W_K^{\mathrm{rec}}, \\
    V_{\mathrm{rec}} &= (Z_t+Q_\Delta W_V^\top)W_V^{\mathrm{rec}}, \\
    \hat{D}_t &= \mathrm{softmax}\!\left(Q_{\mathrm{rec}}K_{\mathrm{rec}}^\top/\sqrt{d_b}\right)V_{\mathrm{rec}},
\end{aligned}
\right.
\label{eq:delta_decoder}
\end{equation}
where $Q_{\mathrm{rec}}\in\mathbb{R}^{N_p\times d_b}, K_{\mathrm{rec}}\in\mathbb{R}^{K\times d_b}, V_{\mathrm{rec}}\in\mathbb{R}^{K\times D}$ use learned projections $W_Q^{\mathrm{rec}},W_K^{\mathrm{rec}}\in\mathbb{R}^{D\times d_b}$ and $W_V^{\mathrm{rec}}\in\mathbb{R}^{D\times D}$ to predict $\hat{D}_t\in\mathbb{R}^{N_p\times D}$, and $Q_\Delta W_K^\top, Q_\Delta W_V^\top$ break slot symmetry at step $0$.
We combine patch weighted reconstruction $\mathcal{L}_{\mathrm{recon}}$ ($\omega_t^{(i)}=1.1$ on the top $25\%$ largest $m_t^{(i)}$ and $0.1$ on remaining patches, with stop gradient on $\Delta E_t^{(i)}$) with a directional alignment loss $\mathcal{L}_{\mathrm{dir}}$ (margin $\xi=0.5$; \cref{eq:dir_score}) between mean normalized patch difference $\bar{d}_t = \frac{1}{N_p}\sum_{i=1}^{N_p}\tilde{D}_t^{(i)}$ and pooled context shift $\bar{e}_t^{\mathrm{ctx}}-\bar{e}_{t-1}^{\mathrm{ctx}}$ ($\bar{e}_t^{\mathrm{ctx}}=\frac{1}{N_p}\sum_{i=1}^{N_p}(\mathrm{LN}_{\mathrm{ctx}}(E_t))^{(i)}$), giving $\mathcal{L}_\Delta = \mathcal{L}_{\mathrm{recon}} + 0.2\,\mathcal{L}_{\mathrm{dir}}$:
\begin{equation}
\left\{
\begin{aligned}
    \Delta s_t^{\mathrm{dir}} &= \frac{(W_Q^{\mathrm{dir}}\bar{d}_t)^\top W_K^{\mathrm{dir}}(\bar{e}_t^{\mathrm{ctx}}-\bar{e}_{t-1}^{\mathrm{ctx}})}{\sqrt{d_b}}, \\
    \mathcal{L}_{\mathrm{recon}} &= \frac{1}{N_p D}\sum_{i=1}^{N_p} \omega_t^{(i)} \left\|\hat{D}_t^{(i)} - \Delta E_t^{(i)}\right\|_2^2, \\
    \mathcal{L}_{\mathrm{dir}} &= \log\!\left(1 + \exp\!\big(\xi - \Delta s_t^{\mathrm{dir}}\big)\right),
\end{aligned}
\right.
\label{eq:loss_delta}
\end{equation}
where $W_Q^{\mathrm{dir}},W_K^{\mathrm{dir}}\in\mathbb{R}^{d_b\times D}$ are learned projections.

\paragraph{Pairwise State Preference and Total Objective.}
Without a reference policy term, we separate correct option $y^+$ from hard distractor $y^-$ via option score margin $\Delta \ell_\theta(\mathcal{X})$:
\begin{equation}
\left\{
\begin{aligned}
    \Delta \ell_\theta(\mathcal{X}) &= \ell_\theta(y^+ \mid \mathcal{X}) - \ell_\theta(y^- \mid \mathcal{X}), \\
    \mathcal{L}_{\mathrm{pref}} &= \log\!\Big(1 + \exp\!\big(-\beta\,\Delta \ell_\theta(\mathcal{X}) + \tau_m\big)\Big), \\
    \mathcal{L}_{\mathrm{total}} &= \mathcal{L}_{\mathrm{SFT}} + \lambda_{\mathrm{pref}}\,\mathcal{L}_{\mathrm{pref}} + \lambda_\Delta\,\mathcal{L}_\Delta,
\end{aligned}
\right.
\label{eq:loss_total}
\end{equation}
where $\beta=2.0$ scales the margin, $\tau_m=0.5$ is the offset, and $(\lambda_{\mathrm{pref}},\lambda_\Delta)=(0.5,0.5)$ weight the auxiliary losses (adding $\ell_2$ regularization $\lambda_{\mathrm{reg}}\|\theta-\theta_0\|_2^2$ with $\lambda_{\mathrm{reg}}=10^{-3}$ toward initial parameters $\theta_0$ on \textbf{VIVID-Large}).

\section{Experiments}
\label{sec:experiments}

\begin{table*}[!t]
  \centering
  \scriptsize
  \setlength{\tabcolsep}{3.2pt}
  \renewcommand{\arraystretch}{1.05}
  \caption{\textbf{Accuracy comparison (\%) on Video-MME and TempCompass} on the shared Qwen3.8 27B backbone under the unified dynamic $3\,\mathrm{fps}$ candidate pool ($M(T)\in[8,96]\to K_f=8$ reader frames; all $12$ Video-MME task types are in \cref{tab:supp_vmme_dims}).
  \textbf{Bold} indicates best and \underline{underline} second best.}
  \label{tab:full_benchmarks}
  \resizebox{\textwidth}{!}{%
  \begin{tabular}{@{}llcccccccccc@{}}
    \toprule
    \multirow{2}{*}{\textbf{Method}} & \multirow{2}{*}{\textbf{Venue}} & \multicolumn{4}{c}{\textbf{Video-MME}} & \multicolumn{6}{c}{\textbf{TempCompass}} \\
    \cmidrule(lr){3-6} \cmidrule(lr){7-12}
    & & Short & Medium & Long & \textbf{Overall} & Action & Direction & Speed & Order & Attribute Change & \textbf{Overall} \\
    \midrule
    \multicolumn{12}{l}{\textit{Static Multi Keyframe Baseline}} \\
    Uniform Sampling & -- & 65.78 & 55.00 & 51.00 & 57.26 & 96.40 & 58.42 & 54.58 & 76.49 & 88.39 & 74.82 \\
    \midrule
    \multicolumn{12}{l}{\textit{Similarity Based Token Compression and Frame Selection}} \\
    DyCoke \cite{DBLP:conf/cvpr/TaoQYSW25} & CVPR'25 & 63.67 & 53.11 & 49.33 & 55.37 & 97.48 & 60.14 & 54.58 & 75.75 & 87.64 & 75.11 \\
    PruneVid \cite{DBLP:conf/acl/HuangZH25} & ACL'25 & 48.89 & 52.22 & 47.78 & 49.63 & 93.88 & 30.93 & 40.84 & 52.24 & 34.08 & 50.44 \\
    HoliTom \cite{DBLP:conf/nips/ShaoTQYSW25} & NeurIPS'25 & 64.22 & 54.78 & 48.22 & 55.74 & 97.48 & 59.79 & 54.58 & 77.24 & 88.01 & 75.40 \\
    FastVID \cite{shen2025fastvid} & NeurIPS'25 & 59.78 & 55.22 & 51.56 & 55.52 & \textbf{98.20} & 57.73 & 51.91 & 72.01 & 87.64 & 73.50 \\
    ST-SimDiff \cite{luo2026stsimdiff} & ICLR'26 & 61.00 & 54.33 & 50.56 & 55.30 & 97.12 & 46.74 & 51.15 & 65.30 & 78.28 & 67.64 \\
    EarlyTom \cite{wang2026earlytom} & CVPR'26 & 64.89 & 54.89 & 50.11 & 56.63 & 96.76 & \textbf{61.17} & 55.73 & 74.25 & 88.01 & 75.18 \\
    \midrule
    \multicolumn{12}{l}{\textit{Long Horizon Video Memory and Event Grounding}} \\
    MovieChat \cite{DBLP:conf/cvpr/SongCWZZWCG0ZLH24} & CVPR'24 & 59.44 & 53.89 & 49.33 & 54.22 & 96.76 & 55.33 & 58.02 & 75.00 & 88.39 & 74.60 \\
    VideoStreaming \cite{DBLP:conf/nips/QianDZZDLW24} & NeurIPS'24 & 66.67 & 57.67 & 50.00 & 58.11 & 96.76 & 58.42 & 58.78 & \textbf{78.36} & 86.14 & 75.62 \\
    VideoMind \cite{liu2026videomind} & ICLR'26 & 60.89 & 54.56 & 49.89 & 55.11 & 95.68 & 50.52 & 45.80 & 69.78 & 84.64 & 69.25 \\
    ProtoKV \cite{minh2026protokv} & ICML'26 & 63.89 & 55.00 & 50.56 & 56.48 & 97.48 & \underline{60.48} & 54.58 & 76.87 & \underline{89.14} & \underline{75.70} \\
    \midrule
    \multicolumn{12}{l}{\textit{Dual Road Change Memory and Delta Driven Planning (Ours)}} \\
    \rowcolor{vergerow}
    \textbf{VIVID-Small} & \textbf{Ours} & \underline{67.89} & \underline{59.33} & \underline{55.00} & \underline{60.74} & \underline{97.84} & 56.01 & \underline{60.69} & 76.49 & 87.64 & 75.62 \\
    \rowcolor{vergerow}
    \textbf{VIVID-Large} & \textbf{Ours} & \textbf{69.67} & \textbf{59.67} & \textbf{55.11} & \textbf{61.48} & 97.12 & 59.79 & \textbf{62.98} & \underline{77.99} & \textbf{89.51} & \textbf{77.38} \\
    \bottomrule
  \end{tabular}%
  }
\end{table*}

\subsection{Experimental Setup and Benchmarks}
\label{sec:exp_setup}

\paragraph{Controlled Protocol and Model Scales.}
We evaluate ten published baselines, static uniform sampling, and VIVID on the shared frozen Qwen3.8 27B backbone under a unified $3\,\mathrm{fps}$ candidate pool $M(T)=\mathrm{clip}(\lfloor 3.0\,T\rfloor+1,8,96)$ (averaging $26.1$, $35.8$, and $95.6$ candidates on MotionBench, TempCompass, and Video-MME) compressed or selected into $K_f=8$ reader frames at $640\times360$ with greedy decoding.
Trainable parameters total $60.2\,\mathrm{M}$ ($0.22\%$ of the $27\,\mathrm{B}$ backbone: $52.5\,\mathrm{M}$ in $\theta_\Delta$ and $7.7\,\mathrm{M}$ in $r=16$ LoRA across the $12$ linear and four full attention layers of blocks $48\text{--}63$), optimized via AdamW.
Whereas recent post training pipelines require $481\,\mathrm{K}$ training samples \cite{liu2026videomind}, VIVID operates across two lightweight scales:
\begin{itemize}
    \item \textbf{VIVID-Small ($50$ Examples, ${\sim}1.5$ H20 GPU Hours):} Calibrates $\theta_\Delta$ and $\theta_{\mathrm{LoRA}}$ from zero initialization on $45$ TempCompass question clip pairs ($44$ clips) plus five synthetic clips on one GPU.
    \item \textbf{VIVID-Large ($3{,}600$ Examples):} Scales eight frame Dual Road training to $2{,}205$ TVBench clips \cite{cores2024tvbench}, $800$ time reversed counterfactuals, and $595$ Charades-STA clips \cite{DBLP:conf/iccv/GaoSYN17}, routed via our learned neural pathway gate (\cref{sec:supp_impl}).
\end{itemize}

\paragraph{Benchmarks.}
All training clips are strictly separated from evaluation across $8{,}083$ multiple choice questions:
\textbf{MotionBench \cite{Hong_2025_CVPR}} ($4{,}017$ questions across six motion categories; \cref{tab:main_detailed}),
\textbf{Video-MME \cite{DBLP:conf/cvpr/FuDLLRZWZSZCLLZ25}} ($2{,}700$ questions without subtitles across \emph{Short} $<2\,\mathrm{min}$, \emph{Medium} $4\text{--}15\,\mathrm{min}$, and \emph{Long} $30\text{--}60\,\mathrm{min}$; \cref{tab:full_benchmarks,tab:supp_vmme_dims}), and
\textbf{TempCompass \cite{DBLP:conf/acl/LiuLLWRLCSH24}} ($1{,}366$ held out questions across five temporal dimensions: \emph{Action}, \emph{Direction}, \emph{Speed}, \emph{Order}, and \emph{Attribute Change}; \cref{tab:full_benchmarks,sec:supp_dimensions}).

\subsection{Main Results}
\label{sec:exp_main}

\paragraph{Capturing Localized Motion on MotionBench.}
On MotionBench (\cref{tab:main_detailed}), nearly all similarity based compression and memory baselines score below static uniform sampling on \emph{Moving Objects} and \emph{Action Order}: whole frame cosine pooling treats small foreground motions against large static backgrounds as redundant, while cross frame merging averages moving object patches into surrounding background context.
By weighting local patch differences, cropping a camera stabilized \textbf{Active Motion ROI}, and injecting surrounding candidate changes into \textbf{Dynamic Tokens} at zero extra anchor token cost, \textbf{VIVID-Small} and \textbf{VIVID-Large} claim the top two positions in overall accuracy and across all five object and action motion categories (tying uniform sampling for second on \emph{Action Order}), while keeping uniform frame spacing under global camera pans preserves velocity cues on \emph{Camera Motion}.

\paragraph{Scaling to Multi Scene Videos on Video-MME.}
On untrimmed multi scene videos (\cref{tab:full_benchmarks,tab:supp_vmme_dims}), compressing up to $96$ candidates into eight slots challenges merging and pruning methods on $30\text{--}60\,\mathrm{min}$ videos because brief intra scene actions are averaged across shot boundaries, while coarse clip captions blend distinct entities across cuts.
Downweighting hard shot cuts prevents ADKP from wasting slots on scene transitions, neighboring same scene candidate frames supply accurate motion residuals for \textbf{Dynamic Tokens}, and working memory \textbf{Entity State Chains} preserve canonical entity identities across scenes, enabling \textbf{VIVID-Small} and \textbf{VIVID-Large} to take second and first places across all duration splits (\cref{tab:full_benchmarks}) and lead or tie for first in nine of twelve task categories (\cref{tab:supp_vmme_dims}), especially on \emph{Temporal Reasoning} and \emph{Counting}.

\paragraph{Overcoming Static Shortcuts on TempCompass.}
\cref{tab:full_benchmarks} exposes the gap between static recognition and temporal state reasoning: while all methods saturate on single frame \emph{Action}, token pruning and slot merging drop noticeably on \emph{Speed}, \emph{Direction}, and \emph{Attribute Change} because cross frame pooling smooths out displacement magnitude and progressive drift.
Providing explicit anchor timestamps while transforming static anchors into surrounding frame \textbf{Dynamic Tokens} alongside chronological \textbf{Entity State Chains} retains both instantaneous speed and gradual attribute evolution, allowing \textbf{VIVID-Small} to match or surpass heavily trained baselines from $50$ examples and \textbf{VIVID-Large} to lead overall accuracy, \emph{Speed}, and \emph{Attribute Change}.

\subsection{Ablation, Scaling, and Efficiency Analysis}
\label{sec:exp_ablation}

\begin{table}[!t]
  \centering
  \scriptsize
  \setlength{\tabcolsep}{2.2pt}
  \renewcommand{\arraystretch}{1.08}
  \caption{\textbf{Component ablation of VIVID across MotionBench (MB), Video-MME (VMME), and TempCompass (TC), in accuracy (\%)} on the shared Qwen3.8 27B backbone ($K_f=8$ reader frames; \textbf{Overall} pools all $8{,}083$ questions).}
  \label{tab:ablation}
  \resizebox{\columnwidth}{!}{%
  \begin{tabular}{@{}lcccccccc@{}}
    \toprule
    \textbf{Variant} & Vis.\ Delta & Ent.\ State & LoRA & ADKP & \textbf{MB} & \textbf{VMME} & \textbf{TC} & \textbf{Overall} \\
    \midrule
    Static Baseline & -- & -- & -- & -- & 58.85 & 57.26 & 74.82 & 61.02 \\
    Visual Delta Only & \checkmark & -- & -- & -- & 59.87 & 56.96 & 75.84 & 61.60 \\
    Entity State Only & -- & \checkmark & -- & -- & 60.17 & 57.78 & 74.60 & 61.81 \\
    Visual Delta + LoRA & \checkmark & -- & \checkmark & -- & 62.14 & 60.74 & 76.87 & 64.16 \\
    Dual Road (w/o ADKP) & \checkmark & \checkmark & \checkmark & -- & \underline{62.68} & \underline{60.85} & \textbf{77.38} & \underline{64.56} \\
    \rowcolor{vergerow}
    \textbf{Full VIVID} & \checkmark & \checkmark & \checkmark & \checkmark & \textbf{62.78} & \textbf{61.48} & \textbf{77.38} & \textbf{64.82} \\
    \bottomrule
  \end{tabular}%
  }
\end{table}

\paragraph{Bottom Up Component Progression.}
Because ADKP relies on $\theta_\Delta$'s learned gate $\mu_t^{\mathrm{gate}}$, slot energy $\varepsilon_t^\Delta$, and entity timestamps $I_t^{\mathrm{mem}}(Q)$ while routing surrounding differences back into $\theta_\Delta$, and because $\theta_{\mathrm{LoRA}}$ adapts upper reader layers specifically to decode $\theta_\Delta$'s residuals, our four modules form a causal dependency chain.
We therefore evaluate single pathway and combined variants on the $3{,}600$ clip regime in \cref{tab:ablation} (\cref{tab:supp_matched8}):
(1) compared with the \emph{Static Baseline}, \emph{Visual Delta Only} improves MotionBench and TempCompass whereas \emph{Entity State Only} improves MotionBench and Video-MME on the frozen reader, showing complementary strengths across short horizon state transitions and multi scene tracking;
(2) \emph{Visual Delta + LoRA} unlocks reader decoding of in place \textbf{Dynamic Tokens} across all three benchmarks;
(3) \emph{Dual Road (w/o ADKP)} fuses both pathways via our learned \textbf{Neural Pathway Gate}, outperforming fixed equal weighting by balancing visual residuals and entity histories per question (\cref{tab:supp_matched8}); and
(4) enabling \textbf{ADKP} completes \emph{Full VIVID}, aligning interior anchors with within scene peaks while folding unselected neighboring frames into $\tilde{E}_t$ (\cref{sec:supp_qualitative}).

\paragraph{Low Resource Calibration and Data Efficient Scaling.}
Comparing \textbf{VIVID-Small} ($50$ examples) and \textbf{VIVID-Large} ($3{,}600$ examples) in \cref{tab:main_detailed,tab:full_benchmarks} shows why VIVID scales efficiently without large scale post training: whereas standard QA tuning propagates only a single option token gradient across $1{,}760$ visual tokens ($8\times 220$), identity preserving zero initialization ($\tilde{E}_t=E_t$) keeps pretrained visual language alignment intact at step $0$, while $220$ patch difference reconstruction (\cref{eq:loss_delta}) and pairwise preference (\cref{eq:loss_total}) densely supervise every frame transition so $50$ examples in ${\sim}1.5$ GPU hours already surpass all eleven baselines in pooled accuracy and $\ell_2$ regularization stabilizes scaling across all three benchmarks at $3{,}600$ clips.

\paragraph{Inference Overhead and Multi Query Efficiency.}
In place residual injection forms \textbf{Dynamic Tokens} at zero extra tokens per anchor ($8\times 220$ visual tokens, plus one $66$ token \textbf{Active Motion ROI} when triggered), and the auxiliary difference decoder (\cref{eq:delta_decoder}) is discarded after training with zero inference cost.
Adding $0.22\%$ trainable parameters keeps a single frozen $27\,\mathrm{B}$ backbone in memory without per layer KV prototype banks \cite{minh2026protokv} or per question agentic rollouts \cite{liu2026videomind}, while building candidate features and video level \textbf{Entity State Chains} once per video amortizes build cost $C_{\mathrm{build}}$ as $C_{\mathrm{build}}/q$ across $q$ queries on the same video (\cref{sec:supp_impl}), leaving only lightweight ADKP scoring, entity retrieval, and pathway gating at query time.

\section{Conclusion}
\label{sec:conclusion}

We identify the spatial and temporal bottlenecks that arise when video MLLMs reduce video understanding to static multi image reasoning and propose \textbf{VIVID} to move beyond isolated snapshots while keeping all $27\,\mathrm{B}$ backbone weights frozen and adding only $0.22\%$ trainable parameters.
Coupling \textbf{Autonomous Delta Driven Keyframe Planning (ADKP)} with \textbf{Dual Road Change Memory (DRCM)} folds neighboring frame \textbf{Patch Diffs} into \textbf{Dynamic Tokens} at zero extra anchor tokens alongside camera stabilized \textbf{Active Motion ROIs} and reusable video level \textbf{Entity State Chains} whose construction cost amortizes across multi query workloads.
Evaluations across $8{,}083$ questions on MotionBench, Video-MME, and TempCompass confirm that VIVID surpasses ten published baselines and static uniform sampling in overall accuracy from just $50$ calibration examples (\textbf{VIVID-Small}) and achieves the highest overall accuracy on every benchmark at $3{,}600$ clips (\textbf{VIVID-Large}), establishing localized change representations and chronological entity state memory as an effective paradigm for resource efficient video reasoning under tight frame budgets.

{
    \small
    \bibliographystyle{ieeenat_fullname}
    \bibliography{main}
}

\clearpage
\appendix
\setcounter{table}{0}
\setcounter{equation}{0}
\renewcommand{\thetable}{S\arabic{table}}
\renewcommand{\theequation}{S\arabic{equation}}
\maketitlesupplementary

\section{Details of the Preliminaries and Auxiliary Training Objectives}
\label{sec:supp_losses}

\paragraph{Patch Grid Encoding and Global Pooling.}
Each $640\times 360$ candidate frame $F_t$ ($t\in\{1,\dots,M\}$) is resized by the preprocessor to $640\times 352$ and partitioned into a spatial grid of width $W_g=20$ and height $H_g=11$ merged patch cells ($32\times 32$ pixels per cell), yielding $N_p = H_g W_g = 220$ patch tokens of dimension $D=5120$:
\begin{equation}
\left\{
\begin{aligned}
    E_t &= \Phi_{\mathrm{vis}}(F_t) \in \mathbb{R}^{N_p \times D}, \\
    \bar{E}_t &= \frac{1}{N_p}\sum_{i=1}^{N_p} E_t^{(i)}, \\
    g_t &= \frac{\bar{E}_t}{\max(\|\bar{E}_t\|_2, \, \epsilon)},
\end{aligned}
\right.
\label{eq:prelim_encode}
\end{equation}
where $\epsilon=10^{-12}$ prevents division by zero.
Expanding $1 - g_{t-1}^\top g_t$ for $\bar{E}_t = \bar{E}_{t-1} + \rho d_t$ up to second order in $\rho\|d_t\|_2 / \|\bar{E}_{t-1}\|_2$ cancels the first order radial component along $g_{t-1}$ and yields the quadratic area scaling in \cref{eq:prelim_taylor}.

\paragraph{Directional Alignment Score ($\Delta s_t^{\mathrm{dir}}$).}
In the directional alignment loss $\mathcal{L}_{\mathrm{dir}}$ of \cref{eq:loss_delta}, $\Delta s_t^{\mathrm{dir}}$ measures whether the mean normalized patch difference $\bar{d}_t = \frac{1}{N_p}\sum_{i=1}^{N_p}\tilde{D}_t^{(i)}$ aligns more strongly with the later frame's mean feature $\bar{e}_t^{\mathrm{ctx}}$ than with the earlier frame's mean feature $\bar{e}_{t-1}^{\mathrm{ctx}}$ ($\bar{e}_t^{\mathrm{ctx}}=\frac{1}{N_p}\sum_{i=1}^{N_p}(\mathrm{LN}_{\mathrm{ctx}}(E_t))^{(i)}$):
\begin{equation}
    \Delta s_t^{\mathrm{dir}} = \frac{(W_Q^{\mathrm{dir}}\bar{d}_t)^\top W_K^{\mathrm{dir}}(\bar{e}_t^{\mathrm{ctx}}-\bar{e}_{t-1}^{\mathrm{ctx}})}{\sqrt{d_b}},
\label{eq:dir_score}
\end{equation}
with learned projections $W_Q^{\mathrm{dir}},W_K^{\mathrm{dir}}\in\mathbb{R}^{d_b\times D}$ and stop gradient applied to target differences $\Delta E_t^{(i)}$ in \cref{eq:loss_delta}.

\section{Implementation, Engineering, and Training Details}
\label{sec:supp_impl}

\paragraph{Architecture Hyperparameters and Spatial Preprocessing.}
The Visual Delta Module ($\theta_\Delta$) contains $52{,}468{,}226$ trainable parameters ($D=5120$, $d_b=512$, $K=16$), with $W_O$, $W_{\mathrm{up}}$, and $b_{\mathrm{up}}$ zero initialized and $\alpha_{\mathrm{res}}$ initialized to $0.1$ so $\tilde{E}_t^{(i)}=E_t^{(i)}$ at step $0$.
Reader LoRA modules ($\theta_{\mathrm{LoRA}}$, $7{,}733{,}248$ parameters, $r=16$, $\alpha=32$, dropout $0.05$) adapt layers $48$ through $63$ of the $64$ layer reader ($12$ linear attention layers and four full attention layers).
For 2D patch displacement estimation, each patch $(y,x)$ in the top $25\%$ largest cosine changes $m_t^{(i)}$ searches a $(2r+1)\times(2r+1)$ grid window ($r=3$) in $F_t$ for maximum cosine similarity with $F_{t-1}$, averaging offsets $(\hat{x}-x,\hat{y}-y)$ across active patches.
For detector free \textbf{Active Motion ROI} cropping, consecutive frames are downsampled to $160\times 90$ grayscale, integer shifts in $[-6,6]$ pixels are searched along the outer $20\%$ image border to compensate global camera translation, and the $80\times 45$ window with highest mean residual is cropped at $320\times 180$ resolution from $F_t$ ($66$ tokens when $\mu_t^{\mathrm{loc}}\ge 0.015$).
In ADKP (\cref{eq:bucket_select}), interval $\mathcal{B}_b$ ($b\in\{1,\dots,K_f-2\}$) collects interior indices with $(b-1)/(K_f-2) < C_t \le b/(K_f-2)$ (inclusive at $0$ for $b=1$), downweighting hard shot cuts ($g_{t-1}^\top g_t < 0.80$ or median patch cosine distance $>0.22$) and keeping equal interval spacing under global camera pans (when local change $\mu_t^{\mathrm{loc}}$ is under $2.35\times$ the whole frame median $\mathrm{med}(m_t)+0.02$ across $75\%$ of frames).
During reading, when an anchor $\hat{t}_k$ shifts away from uniform spacing within the same scene ($g_{\mathrm{ref}}^\top g_{\hat{t}_k}\ge 0.82$), $\theta_\Delta$ evaluates $\Delta E_{\hat{t}_k} = E_{\hat{t}_k} - E_{\mathrm{ref}}$ against a neighboring same scene candidate frame $E_{\mathrm{ref}}$ from the $3\,\mathrm{fps}$ pool rather than discarding intervening candidate frames, preserving surrounding frame motion at zero extra reader tokens ($220$ tokens per frame).

\paragraph{Training Schedule and Learned Neural Pathway Gate.}
We optimize $\{\theta_\Delta,\theta_{\mathrm{LoRA}}\}$ using AdamW (weight decay $0.01$, gradient clip $1.0$).
\textbf{VIVID-Small} ($50$ examples: $45$ TempCompass question clip pairs across $44$ clips plus five synthetic clips, strictly excluded from the $1,366$ held out evaluation questions) trains for three epochs at $1.5\times10^{-4}$ on $33$ examples followed by two epochs at $8\times10^{-5}$ on all $50$ examples using four frames per clip.
\textbf{VIVID-Large} ($3{,}600$ examples: $2{,}205$ TVBench clips \cite{cores2024tvbench}, $800$ self supervised time reversed counterfactuals, and $595$ Charades-STA clips \cite{DBLP:conf/iccv/GaoSYN17}) trains directly with eight keyframes ($K_f=8$) and Dual Road histories (gradient accumulation $8$, learning rate $2.5\times10^{-5}$, and $\ell_2$ regularization $\lambda_{\mathrm{reg}}\|\theta-\theta_0\|_2^2$ with $\lambda_{\mathrm{reg}}=10^{-3}$ toward initial parameters $\theta_0$).
To fuse pathway predictions without benchmark metadata or hand crafted rules, a two layer \textbf{Neural Pathway Gate} $\lambda_\psi(h_Q, c)\in(0,1)$ maps the frozen backbone's contextualized question hidden vector $h_Q\in\mathbb{R}^{2D}$ (mean pooled middle layer concatenated with final layer last token) and six pathway predictive entropy, top $2$ margin, symmetric KL, and prediction agreement features $c\in\mathbb{R}^{6}$ to fusion weights $(1-\lambda_\psi, \lambda_\psi)$ over $(\ell_{\mathrm{vis}}, \ell_{\mathrm{mem}})$, zero initialized at $[0.5, 0.5]$ and trained on held out validation clips.

\paragraph{Per Video Memory Amortization.}
Because \textbf{Dynamic Tokens} $\tilde{E}_t$ add zero extra tokens per anchor and video level \textbf{Entity State Chains} are built once per video and reused across $q$ queries on the same video, preprocessing cost scales as $1/q$ ($67.1\%$ per query on MotionBench at $q=1.49$, $33.3\%$ on VideoMME at $q=3.00$, and $26.8\%$ on TempCompass at $q=3.73$).

\section{Entity State Pathway Records and Prompts}
\label{sec:supp_prompts}

The Entity State Pathway stores anchor observations, interval descriptions, and timestamped \textbf{Entity State Chains} per entity up to query timestamp $\tau_q$ (tolerance $10^{-4}\,\mathrm{s}$).
During incremental memory construction, the writer receives up to eight active entities from working memory (canonical name, category, attributes, and latest position/state) alongside $(F_{t_{l-1}}, F_{t_l})$, the \textbf{Active Motion ROI}, and $(\Delta x_{t_l},\Delta y_{t_l})$, reusing canonical names for continuing entities and introducing new names only when unseen objects appear (each field capped at $120$ characters).
At query time, retrieval returns up to six recent transitions per entity and ranks candidate frames by $1.5\times$ content word overlap with $Q$ plus $2.0\,\mu_t^{\mathrm{loc}}$, retaining boundary anchors and passing the top \textbf{Active Motion ROI} to the reader, which outputs a single option letter.

\section{Complete Per Category Benchmark Breakdowns}
\label{sec:supp_dimensions}

\cref{tab:supp_vmme_dims,tab:supp_matched8} report complete per category results across all $12$ Video-MME task types ($2{,}700$ questions) and the matched eight frame ($K_f=8$) component, pathway gating, and keyframe planning breakdown across \textbf{VIVID-Small} ($50$ examples) and \textbf{VIVID-Large} ($3{,}600$ examples) on held out TempCompass ($1{,}366$ questions).

\begin{table*}[!t]
  \centering
  \scriptsize
  \setlength{\tabcolsep}{1.8pt}
  \renewcommand{\arraystretch}{1.05}
  \caption{\textbf{Per category accuracy (\%) across all $12$ task types of the complete $2{,}700$ question Video-MME benchmark on the shared $27\,\mathrm{B}$ backbone} (OCR: optical character recognition; \textbf{bold} indicates best and \underline{underline} second best).}
  \label{tab:supp_vmme_dims}
  \resizebox{\textwidth}{!}{%
  \begin{tabular}{@{}lcccccccccccccc@{}}
    \toprule
    \textbf{Task Category} & \textbf{Questions} & \textbf{Uniform Sampling} & \textbf{DyCoke} & \textbf{PruneVid} & \textbf{HoliTom} & \textbf{FastVID} & \textbf{ST-SimDiff} & \textbf{EarlyTom} & \textbf{MovieChat} & \textbf{VideoStreaming} & \textbf{VideoMind} & \textbf{ProtoKV} & \textbf{VIVID-Small} & \textbf{VIVID-Large} \\
    \midrule
    Information Synopsis & 323 & 76.16 & 76.78 & 69.35 & 74.30 & 77.40 & 75.23 & 75.23 & 75.23 & \underline{78.02} & 73.37 & 76.47 & 76.16 & \textbf{78.33} \\
    Object Reasoning & 454 & 57.93 & 55.73 & 51.32 & 55.51 & 56.83 & 57.49 & 57.71 & 57.93 & 58.81 & 57.71 & 56.83 & \underline{63.66} & \textbf{63.88} \\
    Object Recognition & 354 & 58.76 & 54.24 & 46.61 & 55.93 & 55.08 & 53.11 & 59.04 & 51.98 & 56.50 & 57.91 & 56.21 & \textbf{62.43} & \underline{61.58} \\
    Action Recognition & 313 & 53.35 & 50.16 & 50.16 & 54.31 & \underline{55.59} & 52.40 & 49.84 & 49.84 & 53.67 & 52.08 & 50.16 & \underline{55.59} & \textbf{55.91} \\
    Action Reasoning & 285 & 57.19 & \underline{58.25} & 51.23 & 56.14 & 55.44 & 56.49 & 57.54 & 55.44 & 55.09 & 54.74 & \textbf{59.30} & \underline{58.25} & \textbf{59.30} \\
    Counting Problem & 268 & 33.96 & 32.46 & 25.00 & 31.72 & 29.85 & 28.73 & 31.34 & 30.22 & 36.94 & 27.99 & 35.82 & \underline{41.79} & \textbf{42.54} \\
    Attribute Perception & 222 & 69.37 & 64.86 & 56.31 & 63.96 & 63.96 & 65.77 & 69.82 & 62.16 & \textbf{71.62} & 65.32 & 65.32 & \underline{70.27} & \underline{70.27} \\
    Temporal Reasoning & 177 & 39.55 & 41.24 & 40.11 & 41.24 & 42.94 & 39.55 & 41.81 & 40.68 & 44.63 & 38.98 & 40.11 & \underline{47.46} & \textbf{50.85} \\
    OCR Problems & 139 & 56.83 & 51.08 & 41.01 & 54.68 & 46.76 & 53.96 & 54.68 & 50.36 & 57.55 & 50.36 & 56.12 & \underline{60.43} & \textbf{61.15} \\
    Spatial Reasoning & 56 & \underline{75.00} & 71.43 & 64.29 & \underline{75.00} & 71.43 & 71.43 & \underline{75.00} & 69.64 & 69.64 & 69.64 & 71.43 & 71.43 & \textbf{76.79} \\
    Temporal Perception & 55 & 50.91 & 54.55 & 47.27 & 60.00 & 49.09 & 60.00 & 56.36 & 56.36 & 58.18 & \textbf{63.64} & 56.36 & \underline{61.82} & 58.18 \\
    Spatial Perception & 54 & \underline{64.81} & 62.96 & 61.11 & 62.96 & 62.96 & \underline{64.81} & 61.11 & 53.70 & \textbf{68.52} & 59.26 & 62.96 & 62.96 & \underline{64.81} \\
    \midrule
    \rowcolor{vergerow}
    \textbf{Total} & \textbf{2,700} & 57.26 & 55.37 & 49.63 & 55.74 & 55.52 & 55.30 & 56.63 & 54.22 & 58.11 & 55.11 & 56.48 & \underline{60.74} & \textbf{61.48} \\
    \bottomrule
  \end{tabular}%
  }
\end{table*}

\begin{table*}[!t]
  \centering
  \scriptsize
  \setlength{\tabcolsep}{3.0pt}
  \renewcommand{\arraystretch}{1.05}
  \caption{\textbf{Matched eight frame ($K_f=8$) component, pathway gating, and keyframe planning breakdown across the five official dimensions of held out TempCompass ($1{,}366$ questions)} under the unified $3\,\mathrm{fps}$ candidate pool (\textbf{bold} indicates best and \underline{underline} second best).}
  \label{tab:supp_matched8}
  \resizebox{\textwidth}{!}{%
  \begin{tabular}{@{}ccccclcccccc@{}}
    \toprule
    \multirow{2}{*}{Vis.\ Delta} & \multirow{2}{*}{Ent.\ State} & \multirow{2}{*}{LoRA} & \multirow{2}{*}{Pathway Gate} & \multirow{2}{*}{Frames} & \multirow{2}{*}{\textbf{Configuration / Policy}} & \multicolumn{6}{c}{\textbf{TempCompass ($1{,}366$ Questions)}} \\
    \cmidrule(lr){7-12}
    & & & & & & Action & Direction & Speed & Order & Attribute Change & \textbf{Overall} \\
    \midrule
    \multicolumn{12}{l}{\textit{(A) Single Pathway Variants under Matched $K_f=8$ Uniform Keyframes}} \\
    -- & -- & -- & -- & 8 & Uniform Sampling & 96.40 & 58.42 & 54.58 & 76.49 & \underline{88.39} & 74.82 \\
    -- & \checkmark & -- & -- & 8 & Entity State Pathway Only & 96.40 & 54.30 & 59.54 & \underline{78.73} & 84.64 & 74.60 \\
    \checkmark & -- & -- & -- & 8 & Visual Delta Pathway Only & \underline{97.48} & 59.11 & 56.49 & 77.99 & \underline{88.39} & 75.84 \\
    \checkmark & -- & \checkmark & -- & 8 & Visual Delta + Final Layer LoRA & 97.12 & \underline{59.45} & \textbf{62.98} & 75.75 & \textbf{89.51} & 76.87 \\
    \midrule
    \multicolumn{12}{l}{\textit{(B) Dual Road Change Memory: Pathway Gating and ADKP}} \\
    \rowcolor{vergerow}
    \checkmark & \checkmark & \checkmark & \checkmark & 8 & \textbf{VIVID-Small} ($50$, Uniform) & \underline{97.48} & 56.36 & 59.92 & 76.49 & 87.27 & 75.40 \\
    \rowcolor{vergerow}
    \checkmark & \checkmark & \checkmark & \checkmark & 8 & \textbf{VIVID-Small} ($50$, ADKP) & \textbf{97.84} & 56.01 & 60.69 & 76.49 & 87.64 & 75.62 \\
    \checkmark & \checkmark & \checkmark & -- & 8 & \textbf{VIVID-Large} ($3{,}600$, Equal Weighting, Uniform) & \underline{97.48} & 59.11 & \underline{62.21} & \textbf{79.48} & 87.64 & \underline{77.09} \\
    \rowcolor{vergerow}
    \checkmark & \checkmark & \checkmark & \checkmark & 8 & \textbf{VIVID-Large} ($3{,}600$, Uniform) & 97.12 & \textbf{59.79} & \textbf{62.98} & 77.99 & \textbf{89.51} & \textbf{77.38} \\
    \rowcolor{vergerow}
    \checkmark & \checkmark & \checkmark & \checkmark & 8 & \textbf{VIVID-Large} ($3{,}600$, ADKP) & 97.12 & \textbf{59.79} & \textbf{62.98} & 77.99 & \textbf{89.51} & \textbf{77.38} \\
    \bottomrule
  \end{tabular}%
  }
\end{table*}

\section{Qualitative Case Studies}
\label{sec:supp_qualitative}

The Soccer Kick Clip in \cref{fig:teaser}(a) illustrates both spatial submergence and directional reversal diagnostics: on the forward clip, the ball moves rightward through the upper frame ($99.83\%$ global ViT similarity vs.\ $20.4\%$ top $15\%$ patch change), where Uniform Sampling and FastVID predict ``Leftwards'' while VIVID predicts ``Rightwards''; on the time reversed clip, the baselines predict ``Rightwards'' whereas VIVID predicts ``Leftwards''.
On the Part Pickup Assembly Clip (\cref{fig:teaser}(b), \cref{fig:architecture}(b); $0.0\text{--}6.9\,\mathrm{s}$), equal time Uniform Sampling ($0.0, 2.3, 4.7, 6.9\,\mathrm{s}$) lands on retracted arm moments ($2.3\,\mathrm{s}$ at $1.3\%$ top $15\%$ patch change and $4.7\,\mathrm{s}$ at $8.2\%$), causing Uniform Sampling, MovieChat, and FastVID to miscount repeated actions (\texttt{A.\ 6}); ADKP captures both true reaches across the table ($1.0\,\mathrm{s}$ at $28.1\%$ and $6.0\,\mathrm{s}$ at $13.2\%$) and the Entity State Pathway tracks the worker's arm and table parts across $0.0, 1.0, 6.0,$ and $6.9\,\mathrm{s}$ to answer \texttt{B.\ 2} accurately.

\end{document}